\documentclass[runningheads]{llncs}
\usepackage{graphicx}
\usepackage{comment}
\usepackage{amsmath,amssymb} 
\usepackage{color}

\usepackage[utf8]{inputenc} 
\usepackage[T1]{fontenc}    
\usepackage{hyperref}       
\usepackage{url}            
\usepackage{booktabs}       
\usepackage{amsfonts}       
\usepackage{nicefrac}       
\usepackage{microtype}      
\usepackage{multirow}
\usepackage{multicol}
\usepackage{wrapfig}
\usepackage{xcolor}
\usepackage{xspace}
\usepackage{caption}
\usepackage{tablefootnote}
\usepackage{epsfig}
\usepackage{makecell}
\usepackage{adjustbox}
\usepackage{array}
\usepackage{wrapfig}
\usepackage{bbm}
\usepackage{blindtext}
\usepackage{fancyhdr}
\usepackage{xcolor}
\usepackage{subcaption}
\usepackage[normalem]{ulem}
\usepackage[permil]{overpic}

\definecolor{OliveGreen}{RGB}{0,180,0}
\definecolor{Grey}{RGB}{94,94,94}

\newcommand{\myparagraph}[1]{\vspace{0.0em}\noindent\textbf{#1}}
\newcommand\onedot{.\enspace}
 
\def\ie{\emph{i.e}\onedot}

\def\etal{\emph{et al}\onedot}

\newcommand{\id}[1]{\textcolor{Grey}{\textbf{\##1}}}

\newcommand{\new}[1]{#1}

\begin{document}
\pagestyle{headings}
\mainmatter
\def\ECCVSubNumber{2701}  

\title{Unselfie: Translating Selfies to Neutral-pose Portraits in the Wild} 


\titlerunning{Unselfie}
%
\author{Liqian Ma \inst{1} \and Zhe Lin\inst{2} \and Connelly Barnes\inst{2} \and Alexei A Efros\inst{2,3} \and Jingwan Lu\inst{2}}
\authorrunning{L. Ma et al.}
%
\institute{$^{1}$KU Leuven \quad $^{2}$Adobe Research \quad $^{3}$UC Berkeley}
\maketitle

\begin{abstract}
Due to the ubiquity of smartphones, it is popular to take photos of one's self, or ``selfies.'' Such photos are convenient to take, because they do not require specialized equipment or a third-party photographer. However, in selfies, constraints such as human arm length often make the body pose look unnatural.
To address this issue, we introduce {\em unselfie}, a novel photographic transformation that automatically translates a selfie into a neutral-pose portrait. 
To achieve this, we first collect an unpaired dataset, and introduce a way to synthesize paired training data for self-supervised learning. Then, 
to {\em unselfie} a photo, we propose a new three-stage pipeline, where we first find a target neutral pose, inpaint the body texture, and finally refine and composite the person on the background. 
To obtain a suitable target neutral pose, we propose a novel nearest pose search module that makes the reposing task easier and enables the generation of multiple \new{neutral-pose} results among which users can choose the best one they like. 
Qualitative and quantitative evaluations show the superiority of our pipeline over alternatives.
\end{abstract}


\section{Introduction}
\label{sec:intro}
\begin{figure*}[b]
\centering

\includegraphics[width=0.92\linewidth]{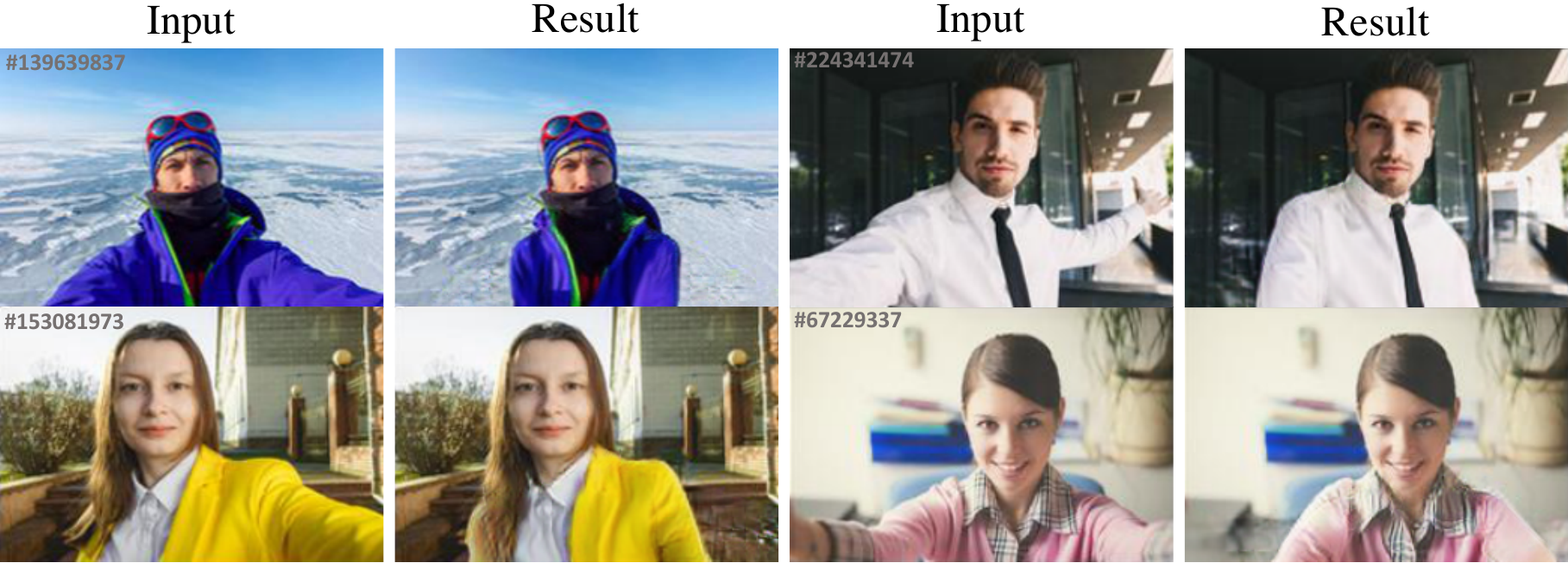}
\caption{We automatically {\em unselfie} selfie photos into neutral-pose portraits. }
\label{fig:teaser}
\end{figure*}

Smartphone cameras have democratized photography by allowing casual users to take high-quality photos. However, there still remains a tension between the ease of capture and the photograph's quality. This is particularly apparent in the case of personal portraits. 
On one hand, it is easy to take a photo of oneself (a selfie) by using the front camera of a smartphone. On the other hand, one can usually take a much higher-quality photograph by relying on an extra photographer, or equipment such as a tripod or selfie stick. 
While less convenient, these avoid the compositional problem that tends to occur in selfies: an unnatural body pose.

In this paper, we introduce a new photographic transformation that we call {\em unselfie}. This transformation aims to make selfie photos look like a well-composed portrait, captured by a third party photographer, showing a neutral body pose with relaxed arms, shoulders and torso.
We call this desired result a ``neutral-pose portrait." The unselfie transform moves any raised arms downward, adjusts the pose of the shoulder and torso, tweaks the details of the clothing and then fills in any exposed background regions (see Figure~\ref{fig:teaser}).

There are three main challenges that we need to tackle in order to be able to {\em unselfie} photos: (1) Paired (selfie, neutral-pose portrait) training data do not exist, so we need to train a model without such data; (2) The same selfie pose can reasonably correspond to multiple plausible neutral poses, so we need to be able to handle this multiplicity; (3) Changing the pose creates holes in the background, so we need to fill in the holes while maintaining a smooth transition between the background and the human body.

We first tried out several previous methods to see if they could address challenge (1). We collected separate sets of selfie and \new{neutral-pose} portraits and used the unsupervised approach CycleGAN~\cite{CycleGAN} for unpaired translation. CycleGAN excels at appearance-level translation that modifies textures and colors, but cannot perform large geometric transformations which are often needed for reposing the complex human body. It also produces unnatural poses with artifacts that result in more noticeable artifacts later in our pipeline. We also tried unsupervised person generation approaches~\cite{DPIG,VUNET}. Though better than CycleGAN, these are not designed for our Unselfie task and produce lower quality results than seen in their papers. As shown in our experiments, these methods result in noticeable artifacts on the generated person images, and texture details are missing because appearance information is compressed heavily. 

Due to these reasons, we instead propose to synthesize (selfie, neutral-pose portrait) pairs and use a self-supervised learning approach. In particular, we propose a way to synthesize paired selfie images from neutral-pose portraits by using a non-parametric nearest pose search module to retrieve the nearest selfie pose given a neutral-pose portrait, and then synthesize a corresponding selfie. We also adopt a nearest pose search module during inference. Given an input selfie pose, we retrieve the best matching neutral poses, which we use to synthesize the final portraits. This addresses challenge (2) by enabling diverse outputs to be synthesized and allowing users to choose among them. 

The synthesized paired data mentioned above can be directly used to train a supervised person image generation network like~\cite{PG2,DSC2018,poseTransfer}, but there still exist noticeable artifacts in the results as shown in our experiments. 
These methods are sensitive to the pixel-level domain gap between our synthetic paired training data and the real selfies testing data (see Fig.~\ref{fig:synthesized_pair_data}).
Inspired by~\cite{CBI_coordInpaint}, we use the coordinate-based inpainting method to inpaint the body texture in UV space. This space is mostly invariant to the original body pose, and is therefore more robust to imperfections in the synthesized data. Additionally, the coordinate-based inpainting method can reuse visible pixels and thus give sharper results.

\begin{figure*} [t]
\scriptsize
  \centering
  \includegraphics[width=1\linewidth]{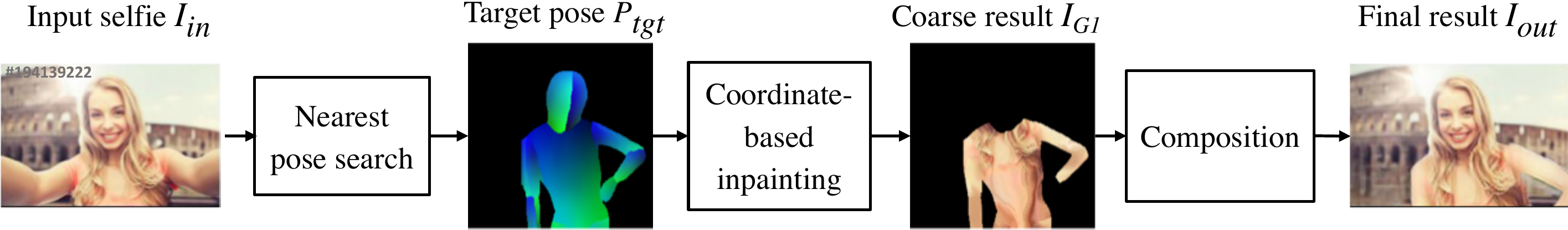}\\
\vspace{-2mm}
\caption{Our three-stage pipeline. Based on the input selfie $I_{in}$, we extract its pose information using DensePose~\cite{densepose}. We perform nearest neighbour search on the pose representation to find the target neutral pose $P_{tgt}$ that has the most similar pose configuration in the upper torso region. Using DensePose, we map the pixels in the input selfie to the visible regions of the target pose and then use coordinate-based inpainting~\cite{CBI_coordInpaint} to synthesize a coarse human body. We then use a composition step to refine the coarse result by adding more details and composite it into the original background. 
}
\label{fig:pipeline}
\end{figure*}

To address challenge (3), we adopt a gated convolutional layer~\cite{yu2019free} based composition network to jointly refine the body appearance, fill the background holes, and maintain smooth transitions between the human body and background. 

Overall, to address the unselfie task,
we propose a three-stage pipeline shown in Fig.~\ref{fig:pipeline}: we first search for a nearest neutral pose in a database, then perform coordinate-based inpainting of the body texture, and finally use a composition module to refine the result and composite it on the background. We conducted several experiments and compared our method with alternatives to demonstrate the effectiveness of our pipeline. 

To the best of our knowledge, this work is the first to target the problem of selfie to neutral-pose portrait translation, which could be a useful and popular application among casual photographers. 
Our contributions include:
1) We collect a new dataset of unpaired selfies and neutral-pose portraits and introduce a way to synthesize paired training data for self-supervised learning; 
2) We introduce a three-stage pipeline to translate selfies into neutral-pose portraits; 
3) We propose a novel nearest pose search module to obtain suitable multi-modal target neutral poses;
4) We design an all-in-one composition module to refine the foreground, complete the background, and compose them together seamlessly.
\section{Related work}
\label{sec:related_work}

\myparagraph{Image Generation and Translation.}
Generative models, such as VAEs~\cite{VAE} and GANs~\cite{GAN,DCGAN,LSGAN,WGAN,bigGAN,styleGAN} can synthesize realistic-looking images from noise. 
To allow more control during the generation process, much research has been devoted to conditional image generation, with class labels~\cite{cGAN}, attributes~\cite{starGAN}, text~\cite{Zhang-stackGAN,Reed-ICML16}, key-points~\cite{reed2016learning,DPIG} and images~\cite{Pix2pix,CycleGAN} as conditioning signals.
Image-to-image translation networks are conditioned on images, such as semantic segmentation maps, edge maps and RGB images~\cite{Pix2pix}. To alleviate the need of collecting paired data, researchers have introduced unsupervised methods based on the ideas of cycle-consistency~\cite{CycleGAN} and shared latent space~\cite{UNIT}. Subsequent works~\cite{lee2018diverse,MUNIT,ma2018exemplar} further extended the unsupervised approach to solve multi-modal image-to-image translation problems by disentangling content and style representations.
These methods mainly focus on appearance manipulation.
Recently, research efforts have also been extended to modify the underlying geometry in image translation tasks~\cite{wu2019transgaga,qian2019make}.
In general, unsupervised image manipulation is quite challenging~\cite{liang2018generative}, especially if the goal is to modify the underlying geometry.

\myparagraph{Image Completion.}
The goal of image completion is to fill in missing regions of an image. Applications of image completion include image inpainting and image out-painting. Traditional patch-based texture synthesis approaches~\cite{barnes2009patchmatch,xu2010image,darabi2012image,huang2014image} work well for images containing simple and repetitive structures, but may fail to handle images of complex scenes. 
To tackle this problem, modern approaches apply deep learning for image completion due to its ability to gain semantic understanding of the image content~\cite{pathak2016context,yu2018generative,yan2018shift,nazeri2019edgeconnect,yu2019free}. 
One of the first deep learning methods for image inpainting is context encoder~\cite{pathak2016context}, which uses an encoder-decoder architecture. 
In order to achieve better results, some prior works apply the PatchMatch idea~\cite{barnes2009patchmatch} at the feature level, such as Contextual Attention~\cite{yu2018generative} and ShiftNet~\cite{yan2018shift}.
Recently, Yu~\etal~\cite{yu2019free} introduce a gated convolutional layer, which learns a dynamic feature gating mechanism for each channel and each spatial location. 
In our framework, we also use the gated convolutional layer~\cite{yu2019free} in both the coordinate-based inpainting network and the composition network to fill in holes in the UV map and the background.

\myparagraph{Person Image Generation.}
Person image generation is a challenging task, as human bodies have complex non-rigid structures with many degrees of freedom~\cite{moeslund2006survey}, 
Previous works in this space usually generate person images by conditioning on these structures. 
Ma~\etal\cite{PG2} propose to condition on image and pose keypoints to transfer the human pose.
\cite{Lassner17ClothNet} and~\cite{dong2018soft} generate clothed person by conditioning on fine-grained body and clothing segmentation. 
\new{Recent works~\cite{men2020controllable,weng2020misc} also extend the conditions to attributes.}
To model the correspondences between two human poses explicitly, recent works introduce flow-based techniques which improve the appearance transfer result quality significantly ~\cite{DSC2018,poseTransfer,DPT,CBI_coordInpaint,han2019clothflow,liu2019liquid,ren2020deep}.
Siarohin~\etal\cite{DSC2018} propose deformable skip-connections to warp the feature maps with affine transformations. 
Grigorev~\etal\cite{CBI_coordInpaint} propose to inpaint the body texture in the UV space which is mostly invariant to the pose in the image space. 
Although these methods achieve good results, they need paired training data, \ie two images containing the same individual in two different body poses, which may be difficult to collect in many applications.
To address this, several unpaired methods have been proposed~\cite{DPIG,VUNET,UPIS}. 
\cite{DPIG} and \cite{VUNET} propose to decompose a person image into different factors which are then used to reconstruct the original image. 
Other works also adopt human parsing algorithms to help out on the difficult unpaired setting ~\cite{Swapnet,song2019unsuper}.
Raj~\etal\cite{Swapnet} generate training pairs from a single image via data augmentation. Inspired by~\cite{Swapnet}, we synthesize selfie data from neutral-pose portrait data to construct paired training data. These human synthesis approaches focus on generating realistic human appearance in a relatively simple background environment (\ie fashion or surveillance datasets) given a target pose. 
Our work on the other hand handles selfie photos captured in the wild that contain a wide variety of backgrounds, lighting conditions, identities and poses. Compared to fashion photos, the background pixels in selfies are of greater importance.

\section{Our Method}
\label{sec:method}

Our goal is to convert a selfie $I_{in}$ into a neutral-pose portrait $I_{out}$ as shown in Fig.~\ref{fig:pipeline}. We collect separate sets of selfie and portrait photos and synthesize paired training data for self-supervised learning (Sect.\ref{sec:dataset}). Due to the complexity of the problem, we solve it in three stages. 
In the first stage, we use a non-parametric nearest pose search module (Sect.\ref{sec:pose_search}) to find the target neutral pose $P_{tgt}$ that closely matches the pose in the upper torso region of the selfie. We then map the pixels in the selfie to regions of the target pose based on the correspondences between the two pose representations. This design makes the remaining problem easier, since most of the pixels can be directly borrowed from the input selfie and thus fewer pixels need to be modified by the remaining steps. In the second stage, inspired by the previous work~\cite{CBI_coordInpaint}, we train a coordinate-based inpainting model to synthesize the coarse appearance of the human body (Sect.\ref{sec:coord_inpaint}). In the final stage, we train a composition model to synthesize details and fix artifacts in the body region caused by pose changes, fill in holes in the background, and seamlessly compose the synthesized body onto the original photo (Sect.\ref{sec:composition}).

We use DensePose~\cite{densepose} in all three stages of our {\em unselfie} pipeline. Unlike keypoint based representations~\cite{cao2018openpose}, which predict a limited amount of pose information described by sparse keypoints, DensePose provides a dense UV-map for the entire visible human body. The UV-map is an ideal pose representation for our purposes, because it provides body shape and orientation information, which are useful for the pose search module. Color values in UV space are also relatively invariant to the person's pose, so this also enables the coordinate-based inpainting step to produce sharp results.

\begin{figure} [t]
\centering
\captionsetup[sub]{font=footnotesize}
\captionsetup[subfigure]{justification=centering}

\begin{subfigure}[b]{0.24\linewidth}
\caption*{Ground truth \\portrait image $I_{tgt}$}
\vspace{-1mm}
\includegraphics[width=\linewidth]{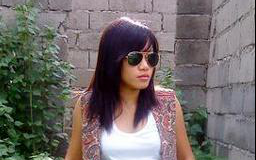}
\end{subfigure}
\begin{subfigure}[b]{0.24\linewidth}
\caption*{Portrait pose \\$P_{tgt}$}
\vspace{-1mm}
\includegraphics[width=\linewidth]{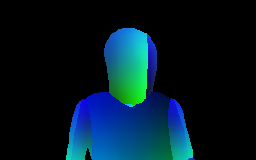}
\end{subfigure}
\begin{subfigure}[b]{0.24\linewidth}
\caption*{\new{Nearest selfie pose $P_{src}$ from $\{P_{selfie}^i\}$}}
\vspace{-1mm}
\includegraphics[width=\linewidth]{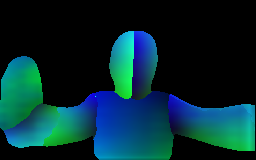}
\end{subfigure}
\begin{subfigure}[b]{0.24\linewidth}
\centering
\caption*{Synthesized selfie image $I_{src}$}
\vspace{-1mm}  
\includegraphics[width=\linewidth]{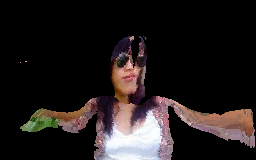}
\end{subfigure}
\caption{Synthesized (portrait, selfie) pairs in image space. Given a neutral-pose portrait $I_{tgt}$, we use DensePose to extract its pose $P_{tgt}$. 
We perform nearest neighbour search to find the closest pose from the selfie pose database. The selfie image $I_{src}$ is synthesized from pixels in $I_{tgt}$ by the correspondence between $P_{tgt}$ and $P_{src}$.
The displayed $I_{tgt}$ and $P_{tgt}$ are cropped due to alignment (Sect.~\ref{sec:pose_search})
}
\label{fig:synthesized_pair_data}
\end{figure} 

\begin{figure} [t]
\centering
\captionsetup[sub]{font=footnotesize}
\captionsetup[subfigure]{justification=centering}

\begin{subfigure}[b]{0.24\linewidth}
\caption*{Ground truth portrait\\texture-map $T_{tgt}$}
\vspace{-1mm}
\includegraphics[width=0.9\linewidth]{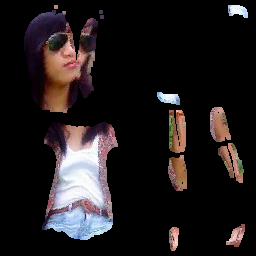}
\end{subfigure}
\begin{subfigure}[b]{0.24\linewidth}
\caption*{Synthesized selfie coordinate-map $C_{src}$}
\vspace{-1mm}
\includegraphics[width=0.9\linewidth]{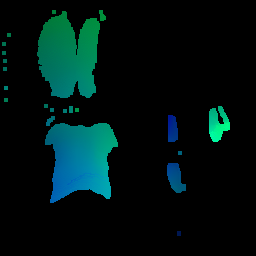}
\end{subfigure}
\begin{subfigure}[b]{0.24\linewidth}
\caption*{Synthesized selfie \\texture-map $T_{src}$}
\vspace{-1mm}
\includegraphics[width=0.9\linewidth]{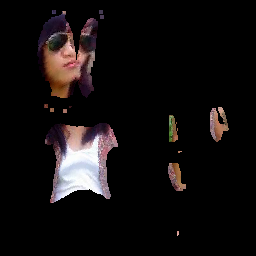}
\end{subfigure}
\caption{Synthesized (portrait, selfie) pairs in UV space to train our coordinate-based inpainting network.}
\label{fig:coord_pair_data}
\end{figure}

\subsection{Datasets}
\label{sec:dataset}
We are not aware of any datasets that contain selfie photos and their corresponding portrait photos taken at the exact same place and time. However, there are many unpaired selfies and neutral-pose portraits online. 

\myparagraph{Unpaired selfies and neutral-pose portraits.}
We collect 23169 photos of people in frontal and neutral poses from the following public datasets: DeepFashion~\cite{DeepFashion}, DeepFashion2~\cite{DeepFashion2}, and ATR~\cite{ATR}.  
We apply the DensePose algorithm to extract the pose information from all the images. 
The extracted DensePose representations form a neutral pose database $\{P_{neutral}^i\}$. 
Because many images in DeepFashion dataset have clean backgrounds, they are not diverse enough for our composition network to learn proper background inpainting. Therefore, we apply a state-of-the-art matting technique~\cite{deepmatting} to extract the foreground humans and paste them into random background images to increase the data diversity. 

We collect 4614 selfie photos from the Internet using the following strategy. We first search with keywords like ``selfie girl,'' ``selfie boy,'' etc. Many photos returned by the search engines contain cellphones. These are not the selfies we desire but are third-person view photos of someone taking selfies. Since Mask R-CNN~\cite{maskrcnn} is pretrained on the COCO dataset~\cite{lin2014microsoft}, which contains person and cell phone classes, we use it to select photos that contain a single person without any cell phones. We then eliminate photos that have disconnected body parts or have any of the frontal upper body parts missing in the DensePose representation. We also use this strategy to create the neutral-pose portrait dataset. Finally, we manually clean up the remaining data in case any of the previous filters fail. We create a 4114/500 split for training and testing. We denote the DensePose representation of selfie photos as $\{P_{selfie}^i\}$. 

\myparagraph{Synthesized paired training data.}
To allow self-supervised training, based on the collected \new{neutral-pose} portraits, we synthesize their corresponding selfie data using DensePose. 
As shown in Fig.~\ref{fig:synthesized_pair_data}, given a neutral pose $P_{tgt}$, we first search for a selfie pose $P_{src}$ from $\{P_{selfie}^i\}$ that matches the input neutral pose the best in the upper torso region. Through DensePose correspondences, we map the portrait image pixels to the nearest selfie pose (see Sect.~\ref{sec:pose_search} for more details). Due to the pose change and self-occlusion, the synthesized selfie might contain holes caused by mistakes of DensePose or by pixels visible in the selfie but not visible in the original portrait such as the armpits or the backside of arms. Though not perfect, these paired images can be used to train supervised human synthesis models like~\cite{PG2,DSC2018,poseTransfer}. However, the results have noticeable artifacts as shown in our experiments in Sect.~\ref{sec:compare}.

Instead of using the synthetic paired images to synthesize pixels directly, we convert them into the UV space (see Fig.~\ref{fig:coord_pair_data}) and perform texture inpainting in the UV coordinate space by building on~\cite{CBI_coordInpaint}. 
\new{In particular, we first obtain ground truth portrait texture map $T_{tgt}$ from $I_{tgt}$ with DensePose mapping. Then, we obtain the selfie coordinate map $C_{src}$ from the nearest selfie pose $P_{src}$ masked by the visible region of $T_{tgt}$. Finally from $C_{src}$, we sample the pixels from $T_{tgt}$ to synthesize the selfie texture map $T_{src}$. $C_{src}$ and $T_{src}$ are used as input to train the coordinate-based inpainting model (see Sect.\ref{sec:coord_inpaint}).}

\subsection{Nearest Pose Search}
\label{sec:pose_search}
Because our goal is to turn a selfie into a neutral-pose portrait, it is important to define what we desire for the target neutral pose. Motivated by the success of retrieval-based image synthesis~\cite{qi2018semi}, we propose a retrieval-based approach where given a selfie pose at testing time, we perform non-parametric nearest pose search to find the neutral poses in the $\{P_{neutral}^i\}$ database that match the input selfie pose the best. 
Compared to pixel-level pose translation using approaches like CycleGAN~\cite{CycleGAN}, our approach has several advantages: (1) It is simpler and more explainable since no training is needed; (2) the retrieved target poses are guaranteed to be natural since they come from real photos; (3) we can generate multiple unselfie results by choosing the top-K most similar poses as the target poses and we can allow users to choose their favorite result; (4) the retrieved poses are similar to the input pose which makes pose correction easier since fewer pixels need to be modified.

\begin{figure}[t]
\scriptsize
  \centering
  \includegraphics[width=1\linewidth]{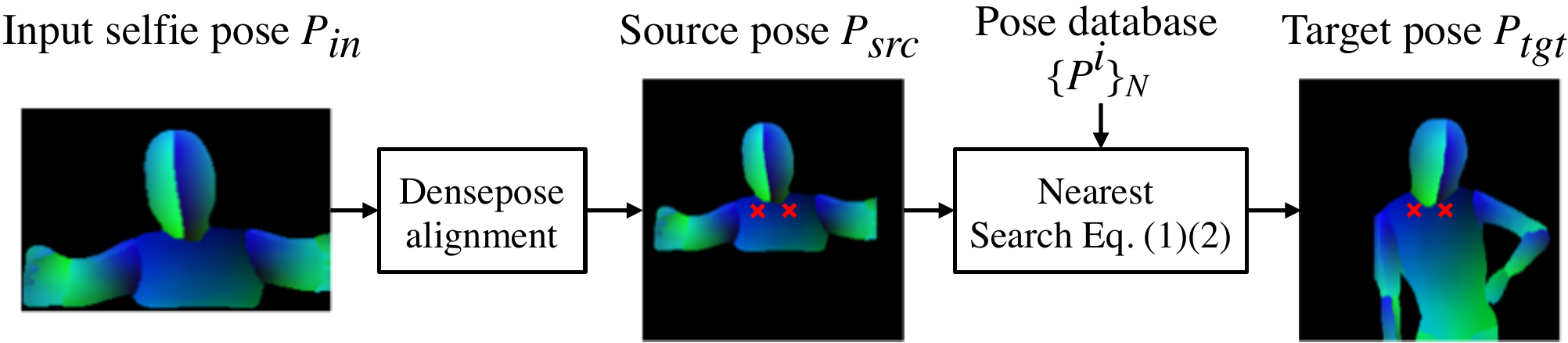}\\
\vspace{-1mm}
\caption{Nearest pose search module. The detected selfie DensePose $P_{in}$ is first aligned so that the two shoulder points are centered in the image $P_{src}$. Then we perform nearest neighbour search to find a target pose $P_{tgt}$ from the neutral pose database that closely matches the input selfie pose in the upper torso region. 
}
\label{fig:pose_search_v3}
\end{figure}

During inference, given an input selfie, we search for the k-nearest target neutral poses. At training time, we reverse the search direction: given a target neutral pose, we search for a matching selfie. This allows us to synthesize synthetic data, which are used to self-supervise the inpainting and composition networks. The procedure is otherwise the same at both training and inference time.

In the remainder of this subsection, we describe the details of our pose search module. As shown in Fig.~\ref{fig:pose_search_v3}, we first align the input selfie pose $P_{in}$ by putting two selected shoulder points in the center of the image to obtain the source pose $P_{src}$.
All neutral poses are also aligned in the same way. We calculate the pose similarity in the frontal torso region excluding the head, since the later stages of our pipeline keep the head region intact and only correct the body pose. The DensePose representation $P$ is an IUV map which contains three channels. The $P^I$ channel contains indices of body parts to which pixels belong, and the $P^{UV}$ channels contain the UV coordinates. 

Based on $P^I$ and $P^{UV}$, we propose a two-step search strategy to calculate pose similarity. First, we search for suitable target poses based on global information such as body shape and position. To determine the global similarity between two poses, we use the following equation: 
\begin{equation}
d^I(P_1,P_2) = \sum\limits_{x\in R_1\cup R_2} \mathbbm{1}(P_1^{I}(x)\neq P_2^{I}(x)),  
\label{eq:d_I} 
\end{equation}
where $R$ refers to the front torso regions of the body. We iterate over all pixels in both torso regions and count the number of pixels that belong to different body parts in the two poses. If there is large body part index mismatch in the torso regions, the two poses are dissimilar at a global level. 

Among the top-K pose candidates selected based on $d_I$, we further improve the ranking by leveraging local pose information given by the UV coordinates. In particular, for pixels belonging to torso regions in both poses, we calculate the sum of the distances of their UV coordinates:
\begin{equation}
    d^{UV}(P_1,P_2) = \sum\limits_{x\in R_1\cap R_2} \|P_1^{UV}(x) - P_2^{UV}(x)\|_2.
    \label{eq:d_UV}
\end{equation}

\subsection{Coordinate-based Inpainting}
\label{sec:coord_inpaint}
\begin{figure*} 
\scriptsize
  \centering
  \includegraphics[width=1\linewidth]{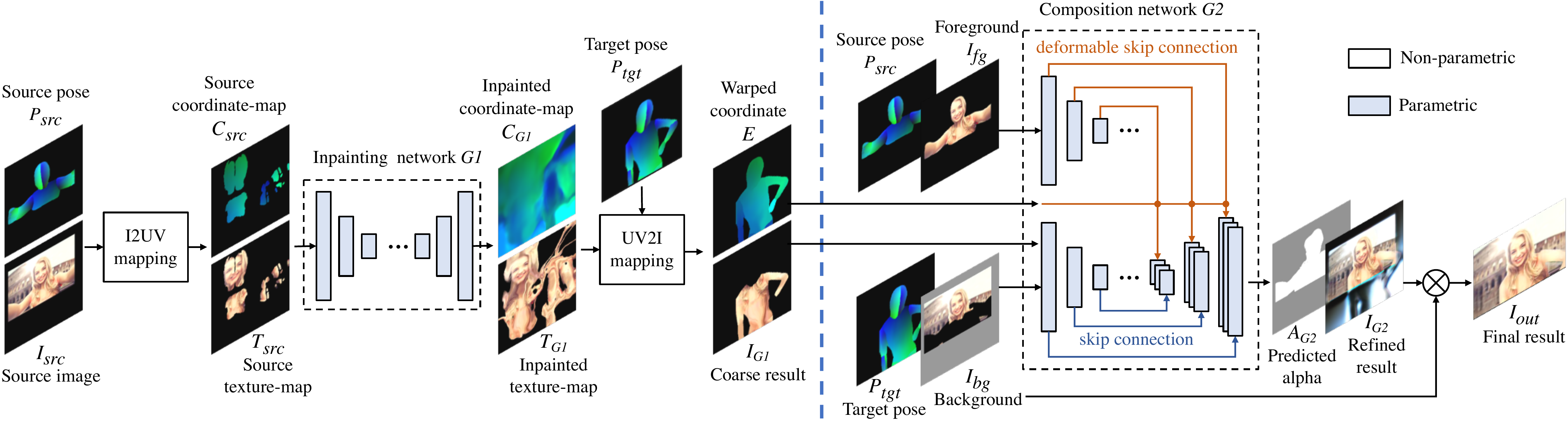}\\
\caption{Left: coordinate-based inpainting stage. Right: composition stage.}
\label{fig:coordInpaint_composition}
\end{figure*}

Inspired by self-supervised image inpainting work~\cite{pathak2016context,yu2019free} and human synthesis work~\cite{CBI_coordInpaint}, we learn to reuse the visible body pixels to fill in the invisible body parts. 
As illustrated in Fig.~\ref{fig:coordInpaint_composition} left, we first use an Image-to-UV (I2UV) mapping to translate pose $P_{src}$ and the color image $I_{src}$ from the image domain to the UV domain.  
Defined in the UV domain, $C_{src}$ stores the associated $\{x,y\}$ coordinates of pixels in the original image space. Likewise in the UV domain, $T_{src}$ contains the RGB colors of the associated pixels in the original image $I_{src}$: these are looked up by using bilinear sampling via $T_{src}=I_{src}(C_{src})$.

After the I2UV mapping, we use an inpainting neural network $G_1$ to inpaint the coordinate-map $C_{src}$. 
We concatenate $C_{src}$ and $T_{src}$ as input to the network.
The network outputs the inpainted coordinate-map $C_{G_1}=G_1(C_{src},T_{src})$. We then perform bilinear sampling to obtain the inpainted texture-map $T_{G_1}=I_{src}(C_{G_1})$.

Finally, we map $C_{G_1}$ and $T_{G_1}$ back to the image space with UV-to-Image (UV2I) mapping using the bilinear sampling operations
$E=C_{G_1}(P_{tgt})$, $I_{G_1}=T_{G_1}(P_{tgt})$. 
%
To train $G_1$, we use three loss functions, identity loss $L_{idt}^{G_1}$, reconstruction loss $L_1^{G_1}$ and perceptual loss $L_P^{G_1}$~\cite{LPIPS} as follows, 
\begin{align}
    L_{idt}^{G_1} = &\mathbb{E}\big[\|C_{G_1}-C_{src}\|_2^2 V_{src} \big],  
    \label{eq:G1_idt_loss}  \\
    L_1^{G_1} = &\mathbb{E}\big[\|T_{G_1}-T_{tgt}\|_1 V_{tgt} \big],
    \label{eq:G1_L1_loss}  \\
    L_P^{G_1} = &\mathbb{E}\big[\|\phi(T_{G_1})-\phi(T_{tgt})\|_2^2 V_{tgt}\big]
    \label{eq:G1_LPIPS_loss},
\end{align}
\new{$V_{src}$ and $V_{tgt}$ are binary masks that select the non-empty regions in the coordinate map $C_{src}$ and $T_{tgt}$, respectively.} $T_{tgt}$ is the ground truth texture mapped from image domain to UV domain (Fig.~\ref{fig:coord_pair_data} left), that is, $T_{tgt}=I2UV(I_{tgt})$. 
The identity loss encourages the existing coordinates to stay unchanged while the network synthesizes coordinates elsewhere. The reconstruction loss and perceptual loss are performed in the pixel space instead of the coordinate space and use the ground truth image for supervision. 
The overall loss for $G_1$ is
\begin{equation}
    \min\limits_{G_1}
    L^{G_1}=L_1^{G_1}+\lambda_{1}L_{P}^{G_1}+\lambda_{2} L_{idt}^{G_1},
    \label{eq:G1_full_loss}  
\end{equation}

\subsection{Composition}
\label{sec:composition}
The advantage of doing inpainting in the coordinate space is that the network can copy and paste original pixels to fill in missing regions based on body symmetry and therefore the synthesized pixels tend to look sharp. However, in some cases, a small number of visible pixels get copied into a much larger region resulting in flat and unrealistic appearance.
In addition, when arms are moved down, holes will appear in the background due to dis-occlusion. 

To address these problems, we use an all-in-one composition network to add details and fix artifacts in the body region and fill in the gaps between the body and the background by synthesizing a natural transition. 
As illustrated in Fig.~\ref{fig:coordInpaint_composition} right, we use a U-net architecture equipped with gated convolutional layers~\cite{yu2019free} for $G_2$. The U-net architecture helps preserve the high resolution features through skip-connections. 
The gated convolutional layer improves the inpainting result quality when dealing with holes of arbitrary shape.
To keep the body appearance more consistent, we also use deformable skip connections~\cite{DSC2018,CBI_coordInpaint} to propagate the appearance information from the source image $I_{src}$ to the result despite large changes in poses. The network synthesizes missing foreground pixels, fills in background holes and also produces an alpha mask $A_{G_2}$. $A_{G_2}$ is used to blend the synthesized pixels $I_{G_2}$ into the background image $I_{bg}$, i.e. $I_{out}=I_{G_2} A_{G_2} + I_{bg}(1-A_{G_2})$.
%
We add the original head and neck regions into $I_{bg}$ so that after blending the head regions will remain untouched.

To train $G_2$, we apply reconstruction loss $L_1^{G_2}$, perceptual loss $L_P^{G_2}$~\cite{LPIPS}, alpha loss $L_A^{G_2}$ and adversarial loss $L_{adv}^{G_2,D}$,
\begin{align}
    &L_1^{G_2} = \mathbb{E}\big[\|I_{out}-I_{tgt}\|_1 (1+H)\big],
    \label{eq:G2_L1_loss}  \\
    &L_P^{G_2} = \mathbb{E}\big[\|\phi(I_{out})-\phi(I_{tgt})\|_2^2(1+H)\big]
    \label{eq:G2_LPIPS_loss},\\
    &L_A^{G_2} = \mathbb{E}\big[\|A_{G_2}-H\|_1\big],
    \label{eq:G2_Aplha_loss}  \\
    &\min\limits_{G_2}\max\limits_{D} 
    L_{adv}^{G_2,D} = {\mathbb{E}}\big[(D(I_{tgt}))^2(1+H)\big] \nonumber \\
    & ~~~~~~ +{\mathbb{E}}\big[(D(I_{out})-1)^2(1+H)\big], 
    \label{eq:G2_GAN_loss}
\end{align}
where $I_{tgt}$ denotes the ground truth neutral-pose portrait. \new{$H \in [0,1]$ is a binary spatial mask to encourage the network to focus more on synthesizing foreground and filling dis-occluded holes and the details are explained later. When applied to different spatial size, $H$ will be resized to the corresponding spatial size by nearest-neighbor scaling accordingly. As to the adversarial learning, we use the same residual discriminator as that of~\cite{poseTransfer}.}
The overall loss for $G_2$ is
\begin{equation}
    \min\limits_{G_2}\max\limits_{D}
    L^{G_2}=\lambda_{3}L_1^{G_2}+\lambda_{4}L_{P}^{G_2}+\lambda_{5} L_{adv}^{G_2,D}+L_{A}^{G_2}.
    \label{eq:G2_full_loss}  
\end{equation}

There is a big domain gap between the training and testing data. During testing, arms in real selfies are moved downward revealing a large hole in the background. During training, we also mimic the dis-occluded background holes.
In particular, we calculate a binary mask $H=H_{selfie}\cup H_{neutral}$, which is also used in Eq.~\ref{eq:G2_L1_loss} to~\ref{eq:G2_GAN_loss}. $H_{selfie}$ and $H_{neutral}$, which are estimated using an off-the-shelf DeepMatting model~\cite{deepmatting} and binarized with threshold $0.1$, denote the body regions from the selfie and the neutral-pose portrait, respectively. The synthesized hole mask $H$ is then applied to $I_{bg}$ to mimic dis-occluded background holes.

\section{Experiments}
\label{sec:experiments}

\new{We compare our approach with several prior work through a qualitative evaluation, a user study and a quantitative evaluation\footnote{More results and implementation details are reported in the supplementary materials.}.} Note that none of the previous approaches address exactly our unselfie problem, so we cannot compare our approach with previous work using their datasets and result quality for previous work on our dataset is worse than the result quality in those papers. 
We present ablation studies to validate the effectiveness of different algorithm choices. Finally, we discuss the limitations and future work. \new{If not otherwise specified, we use the top-1 retrieved neutral pose as target pose.}
Note that there may be JPEG compression artifacts in the input selfies. 

\subsection{Comparisons with existing methods}
\label{sec:compare}
Since we defined a brand new unselfie application, there is no prior work to compare to that addresses the exact same problem. Nevertheless, we introduce some modifications to two state-of-the-art human synthesis methods, DPIG~\cite{DPIG} and PATN~\cite{poseTransfer}, so that we can compare to them directly in our new application setting. 
Note that these methods synthesize pixels based on a pre-specified target pose. To make their approaches work, we need to perform our proposed nearest pose search module to calculate $P_{tgt}$ and then use their approaches to synthesize the final pixels. 
DPIG is a key-points based unsupervised approach. For fair comparison, we replace their key-points with the DensePose representation. We also made various other improvements for it to produce comparable results to ours (see supplementary material). PATN is a key-points based supervised method, so we use our synthesized paired images for self-supervised training \new{by using DensePose IUV map as the input pose representation and feeding $I_{src}$, $I_{bg}$ as input to their model.} 
In the supplementary material, we also compare our approach with another keypoint-based unsupervised approach VUNET~\cite{VUNET}. Due to low result quality and space limitations, we do not show those results here. 
\begin{figure*}[t]
\centering

\begin{subfigure}[b]{.24\linewidth}
\caption*{Input selfie}
\vspace{-1.5mm}
\begin{overpic}[width=\linewidth]{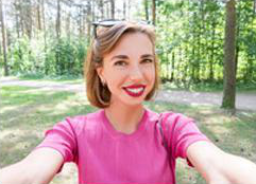}
\put(0,655) {\scalebox{0.65}{\id{212727509}}}
\end{overpic}
\end{subfigure}
\begin{subfigure}[b]{.24\linewidth}
\caption*{DPIG~\cite{DPIG}}
\vspace{-1.5mm}
\includegraphics[width=\linewidth]{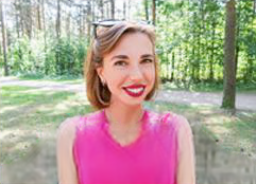}
\end{subfigure}
\begin{subfigure}[b]{.24\linewidth}
\caption*{PATN~\cite{poseTransfer}}
\vspace{-1.5mm}
\includegraphics[width=\linewidth]{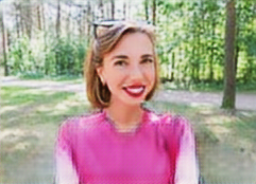}
\end{subfigure}
\begin{subfigure}[b]{.24\linewidth}
\caption*{Ours}
\vspace{-1.5mm}
\includegraphics[width=\linewidth]{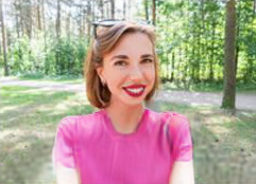}
\end{subfigure}

\begin{subfigure}[b]{.24\linewidth}
\vspace{-1.5mm}
\begin{overpic}[width=\linewidth]{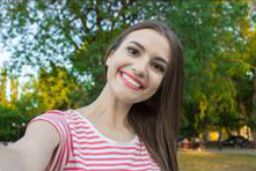}
\put(0,610) {\scalebox{0.65}{\id{168103021}}}
\end{overpic}
\end{subfigure}
\begin{subfigure}[b]{.24\linewidth}
\vspace{-1.5mm}
\includegraphics[width=\linewidth]{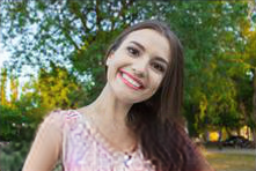}
\end{subfigure}
\begin{subfigure}[b]{.24\linewidth}
\vspace{-1.5mm}
\includegraphics[width=\linewidth]{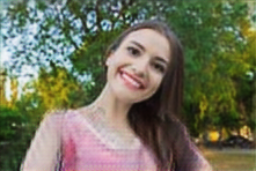}
\end{subfigure}
\begin{subfigure}[b]{.24\linewidth}
\vspace{-1.5mm}
\includegraphics[width=\linewidth]{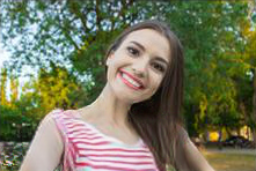}
\end{subfigure}



\begin{subfigure}[b]{.24\linewidth}
\vspace{-1.5mm}
\begin{overpic}[width=\linewidth]{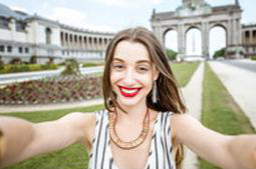}
\put(0,600) {\scalebox{0.65}{\id{162277318}}}
\end{overpic}
\end{subfigure}
\begin{subfigure}[b]{.24\linewidth}
\vspace{-1.5mm}
\includegraphics[width=\linewidth]{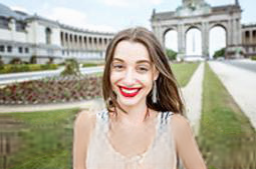}
\end{subfigure}
\begin{subfigure}[b]{.24\linewidth}
\vspace{-1.5mm}
\includegraphics[width=\linewidth]{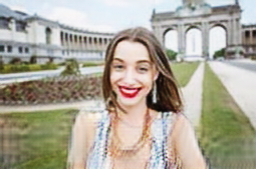}
\end{subfigure}
\begin{subfigure}[b]{.24\linewidth}
\vspace{-1.5mm}
\includegraphics[width=\linewidth]{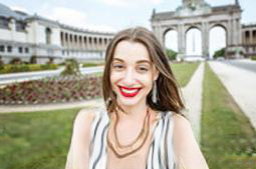}
\end{subfigure}

\begin{subfigure}[b]{.24\linewidth}
\vspace{-1.5mm}
\begin{overpic}[width=\linewidth]{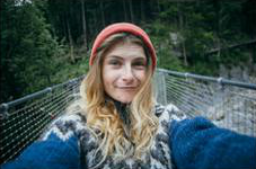}
\put(0,610) {\scalebox{0.65}{\id{225137362}}}
\end{overpic}
\end{subfigure}
\begin{subfigure}[b]{.24\linewidth}
\vspace{-1.5mm}
\includegraphics[width=\linewidth]{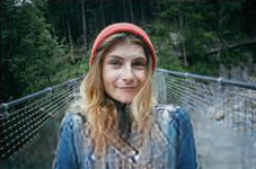}
\end{subfigure}
\begin{subfigure}[b]{.24\linewidth}
\vspace{-1.5mm}
\includegraphics[width=\linewidth]{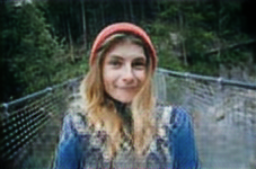}
\end{subfigure}
\begin{subfigure}[b]{.24\linewidth}
\vspace{-1.5mm}
\includegraphics[width=\linewidth]{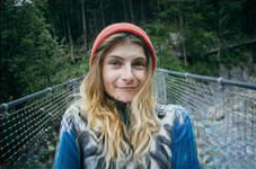}
\end{subfigure}


\caption{Comparisons with state-of-the-art methods. Please zoom in to see details.}
\label{fig:comparison}
\end{figure*}

\myparagraph{Qualitative evaluation.} 
Fig.~\ref{fig:comparison} shows that our method synthesizes more photo-realistic body and background appearance compared to prior art. 
\new{We manually picked the best target pose out of results using our top-5 retrieved poses. The multi-modal results are reported in the supplementary material.}
In the top row, the short-sleeved clothing style is better preserved. In the second row, the stripe pattern is better preserved, and the synthesized arms are sharper. 
\new{In the bottom two rows}, our method synthesizes better clothing and background details and produces more natural transitions between foreground and background. 
The reasons that our method outperforms the baselines are: 1) unsupervised methods, like DPIG, encode images into heavily compressed features, which results in loss of details and texture inconsistency between the generated output and the input. They perform well in more constrained settings (clean backgrounds and simple texture), while our task involves complex images in the wild; 2) these baseline methods are more sensitive to the domain gap between training and testing data since they directly synthesize image pixels. Our method performs foreground inpainting in the coordinate space and then uses a composition module to refine details and fill in background holes and thus is less sensitive to the domain gap between imperfect synthesized selfies at training and perfect selfies at testing. 

\begin{table}[b]
\begin{tabular}{cc}
    \begin{minipage}{.57\linewidth}
    \centering
    \footnotesize
    \begin{tabular}
    {@{\extracolsep{\fill}} l c c c}
    \toprule 
    Model & Human Prefers Ours & FID$\downarrow$ & KID$\downarrow$ \\
    \midrule[0.6pt]	
    	DPIG~\cite{DPIG} & 0.798 & 88.27 & 0.026 \\
    	VUNET~\cite{VUNET} & 0.851 & 135.90 & 0.077 \\
        PATN~\cite{poseTransfer} & 0.822 & 104.74 & 0.041 \\
        Ours & N/A & \textbf{71.93} & \textbf{0.014} \\
    \bottomrule[0.6pt]	
    \end{tabular}
    \caption{User study and, FID/KID scores.
    } 
    \label{tab:user_fid}
    \end{minipage} &
    
    \hspace{3mm}

    \begin{minipage}{.37\linewidth}
    \centering
    \footnotesize
    \begin{tabular}
    {@{\extracolsep{\fill}} l c c}
    \toprule 
    Model & FID$\downarrow$ & KID$\downarrow$ \\
    \midrule[0.6pt]	
    	Ours w/o $L_P^{G_2}$ & 82.09 & 0.019 \\
        Ours w/o Deform & 73.87 & 0.017 \\
    	Ours w/o Gated & 72.89 & \textbf{0.014} \\
        Ours & \textbf{71.93} & \textbf{0.014} \\
    \bottomrule[0.6pt]	
    \end{tabular}
    \caption{Ablation study.} 
    \label{tab:ablation}
    \end{minipage} 
\end{tabular}
\end{table}

\myparagraph{User study.}
For a useful real-world application, we believe qualitative perceptual evaluation is more important. Thus, we perform a user study on Amazon Mechanical Turk (AMT). 
Similar to previous works~\cite{MUNIT,lee2018diverse}, given the input selfie and a pair of results generated by our approach and one of the baseline approaches, users are asked to pick one that looks better than the other. 
Within each Human Intelligence Task (HIT), we compare our method with the same baseline method. We randomly generate 200 result pairs including 10 sanity pairs where the answers are obvious. After filtering out careless users based on their answers on the sanity pairs, we calculate the user study statistics using the remaining 190 questions. We have three HITs for three baseline methods. Each HIT is done by 20 users. 
As shown in Table.~\ref{tab:user_fid}, our method is preferred over others.
We assume a null hypothesis that on average, half the users prefer ours for a given paired comparison. We use a one-sample permutation t-test~\cite{permutation} to measure $p$ using $10^6$ permutations and find $p < 10^{-6}$ for the 3 baselines.

\myparagraph{Quantitative evaluation.}
Since we do not have the ground truth neutral portraits corresponding to input selfies, we cannot use metrics like SSIM. To quantitatively compare our result quality with other baselines, we report Frechet Inception Distance (FID)~\cite{fid} and Kernel Inception Distance (KID)~\cite{KID} as shown in Table.~\ref{tab:user_fid}. 
We aim to translate the body into a neutral pose while keeping the rest of the image intact. Therefore, a good translation method should have low FID and KID values when compared to both the portraits and the selfie domains. 
As suggested by~\cite{mejjati2018unsupervised}, we combine both real selfie and real portrait images into the real domain, and compute the FID and KID values between our synthesized results (\ie fake domain) and the real domain. 
The mean FID and KID values are averaged over 10 different splits of size 50 randomly sampled from each domain. The trend of FID and KID is consistent with the user study result, and our method outperforms others significantly.

\begin{figure}[t]
\centering

\begin{subfigure}[b]{.18\linewidth}
\caption*{Input selfie}
\vspace{-1.5mm}
\begin{overpic}[width=\linewidth]{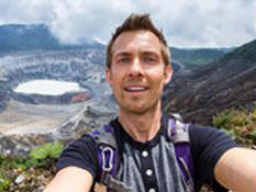}
\put(0,680) {\scalebox{0.6}{\id{120915150}}}
\end{overpic}
\end{subfigure}
\begin{subfigure}[b]{.18\linewidth}
\caption*{$d_I$}
\vspace{-1.5mm}
\includegraphics[width=\linewidth]{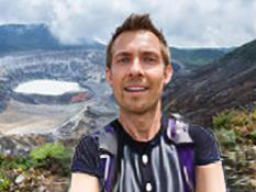}
\end{subfigure}
\begin{subfigure}[b]{.18\linewidth}
\caption*{$d_{UV}$}
\vspace{-1.5mm}
\includegraphics[width=\linewidth]{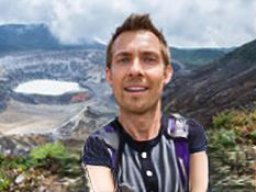}
\end{subfigure}
\begin{subfigure}[b]{.18\linewidth}
\caption*{Ours}
\vspace{-1.5mm}
\includegraphics[width=\linewidth]{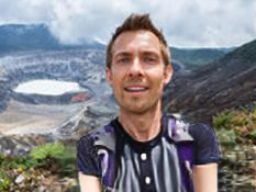}
\end{subfigure}
\caption{Ablation study results for nearest pose search.
}
\label{fig:ablation_pose}
\end{figure}

\begin{figure}[t]
\centering
\begin{subfigure}[b]{.19\linewidth}
\caption*{Input selfie}
\vspace{-1.5mm}
\begin{overpic}[width=\linewidth]{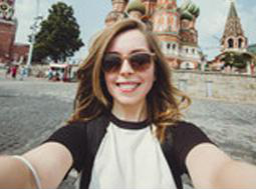}
\put(0,680) {\scalebox{0.6}{\id{133457041}}}
\end{overpic}
\end{subfigure}
\begin{subfigure}[b]{.19\linewidth}
\caption*{w/o $L_P^{G_2}$}
\vspace{-1.5mm}
\includegraphics[width=\linewidth]{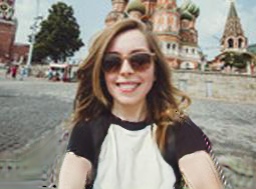}
\end{subfigure}
\begin{subfigure}[b]{.19\linewidth}
\caption*{w/o Deform}
\vspace{-1.5mm}
\includegraphics[width=\linewidth]{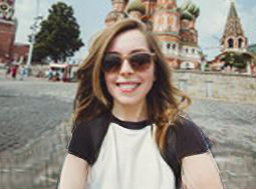}
\end{subfigure}
\begin{subfigure}[b]{.19\linewidth}
\caption*{w/o Gated}
\vspace{-1.5mm}
\includegraphics[width=\linewidth]{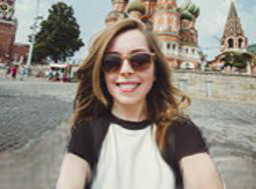}
\end{subfigure}
\begin{subfigure}[b]{.19\linewidth}
\caption*{Ours}
\vspace{-1.5mm}
\includegraphics[width=\linewidth]{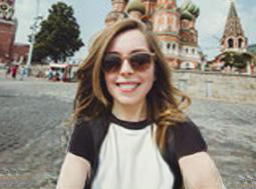}
\end{subfigure}
\caption{Ablation study for composition network.}
\label{fig:ablation_noGated}
\end{figure}

\subsection{Ablation Study}
\myparagraph{Analysis of the pose search module.}
We compare three ablation settings: 1) using body index distance only (\ie $d_I$); 2) using UV distance only (\ie $d_{UV}$); 
3) using our two-step strategy.
As shown in Fig.~\ref{fig:ablation_pose}, the neutral poses retrieved by $d_I$ have a reasonable body shape and size but the local coordinates mismatch resulting in the cloth regions being distorted (see the elongated buttons). 
For the neutral poses retrieved by $d_{UV}$, the body shape and size are not compatible with the head region. 
The retrieved body part is too small. 
Our two-step strategy combines the benefits of $d_I$ and $d_{UV}$ and retrieves better poses. 

\myparagraph{Analysis of the compositing network.} We compare three ablation settings. \textbf{w/o $L_P^{G_2}$:} removing perception loss $L_P^{G_2}$. \textbf{w/o Deform:} removing deformable skip-connection. \textbf{w/o Gated:} use normal conv layer instead of gated conv layer~\cite{yu2019free}. 
As shown in Fig.~\ref{fig:ablation_noGated}, removing any of the components will result in noticeable result degradation. Our full setting synthesizes more details in the foreground, smoother transition between the foreground and the background and also better fills the large background holes. The quantitative results reported in Table.~\ref{tab:ablation} are consistent with the above observations.

\begin{figure}[h]
\centering

\begin{subfigure}[b]{.175\linewidth}
\centering
\begin{overpic}[height=0.64in]{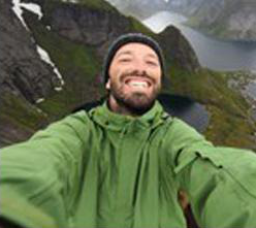}
\put(0,810) {\scalebox{0.65}{\id{109067715}}}
\end{overpic}
\end{subfigure}
\begin{subfigure}[b]{.175\linewidth}
\centering
\includegraphics[height=0.64in]{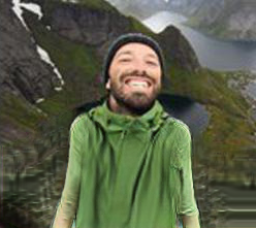}
\end{subfigure}
\hspace{1mm}
\begin{subfigure}[b]{0.20\linewidth}
\centering
\begin{overpic}[height=0.64in]{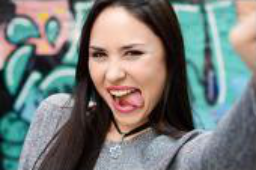}
\put(0,600) {\scalebox{0.65}{\id{121680430}}}
\end{overpic}
\end{subfigure}
\begin{subfigure}[b]{0.20\linewidth}
\centering
\includegraphics[height=0.64in]{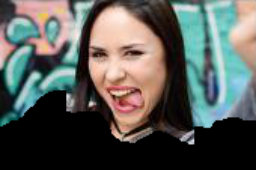}
\end{subfigure}
\begin{subfigure}[b]{0.20\linewidth}
\centering
\includegraphics[height=0.64in]{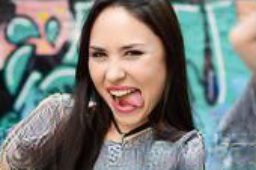}
\end{subfigure}
\caption{Failure cases. Left: input, result. Right: input, foreground mask, result. 
}

\label{fig:failure}
\end{figure}

\begin{figure}
\centering
\captionsetup{font=small,labelfont=normalsize,justification=centering}
\vspace{-1mm}

\begin{subfigure}[b]{0.19\linewidth}
\caption*{Input selfie}
\vspace{-1mm}
\begin{overpic}[width=\linewidth]{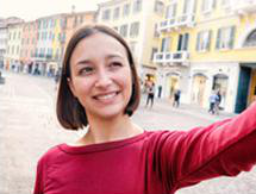}
\put(0,690) {\scalebox{0.65}{\id{206713499}}}
\end{overpic}
\end{subfigure}
\begin{subfigure}[b]{0.19\linewidth}
\caption*{BG $I_{bg}$}
\vspace{-1mm}
\includegraphics[width=1\linewidth]{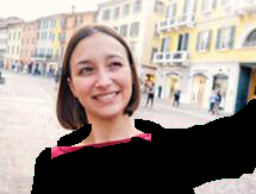}
\end{subfigure}
\begin{subfigure}[b]{0.19\linewidth}
\caption*{Inpainted BG $I_{ibg}$}
\vspace{-1mm}
\includegraphics[width=1\linewidth]{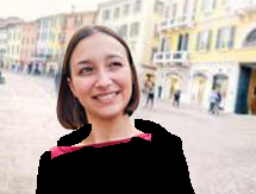}
\end{subfigure}
\begin{subfigure}[b]{0.19\linewidth}
\caption*{Result with $I_{bg}$}
\vspace{-1mm}
\includegraphics[width=1\linewidth]{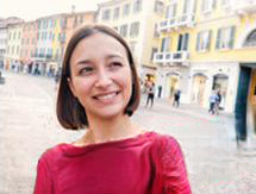}
\end{subfigure}
\begin{subfigure}[b]{0.19\linewidth}
\caption*{Result with $I_{ibg}$}
\vspace{-1mm}
\includegraphics[width=1\linewidth]{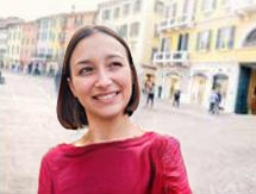}
\end{subfigure}

\captionsetup{font=normalsize}
\caption{Results of using inpaited background.}
\label{fig:inpaintBG}
\end{figure}

\subsection{Limitations}
Our approach has several limitations. 
First, as shown in Fig.~\ref{fig:failure} left, for challenging non-frontal selfie poses/viewpoints, our nearest pose search module might struggle with finding compatible \new{neutral poses}, which results in the synthesized result containing arms or shoulders that are too slim or wide  compared to the head region. This problem happens less than 10\% of the time in our top-1 result, and users can usually find a good compatible pose from our top-5 results. Second, this example also reveals the limitation of our background synthesis.
\new{We also show one example obtained by inpainting the background with an extra off-the-shelf model~\cite{zeng2020high} in Fig.~\ref{fig:inpaintBG} to demonstrate the  benefits from the image inpainting model trained on large-scale datasets.}
Finally, our system is prone to errors in DensePose detection. Fig.~\ref{fig:failure} right, DensePose failed to detect her left arm as foreground. Therefore the composition module retained her left arm in the result.

\section{Conclusion}
\label{sec:conclusion}
In this work, we introduce a novel ``unselfie'' task that translates a selfie into a portrait with a neutral pose. We collect an unpaired dataset and introduce a way to synthesize paired training data for self-supervised learning. We design a three-stage framework to first retrieve a target neutral pose to perform warping, then inpaint the body texture, and finally fill in the background holes and seamlessly compose the foreground into the background. Qualitative and quantitative evaluations demonstrate the superiority of our framework over other alternatives.

\noindent{\textbf{Acknowledgements:}}
This work was partially funded by Adobe Research. We thank He Zhang for helping mask estimation. \new{Selfie photo owners: \#139639837-Baikal360, \#224341474-Drobot Dean, \#153081973-MaximBeykov, \#67229337-Oleg Shelomentsev, \#194139222-Syda Productions, \#212727509-Photocatcher, \#168103021-sosiukin, \#162277318-rh2010, \#225137362-BublikHaus, \#120915150-wollertz, \#133457041-ilovemayorova, \#109067715-Tupungato, \#121680430-Mego-studio, \#206713499-Paolese -- stock.adobe.com.
}

\clearpage

\begin{center} 
\Large{\textbf{Unselfie: Translating Selfies to Neutral-pose Portraits in the Wild -- Supplementary material.}}
\end{center} 

\setcounter{section}{0}
\renewcommand\thesection{\Alph{section}}
\setcounter{figure}{0}
\setcounter{table}{0}
\renewcommand{\thefigure}{S\arabic{figure}}
\renewcommand{\thetable}{S\arabic{table}}

In this supplementary material, we provide additional results, more visual comparisons with prior art and implementation details. 

\section{Results of using off-the-shelf inpainting network for background}
\begin{figure}[h]
\scriptsize
  \centering
  \includegraphics[width=1\linewidth]{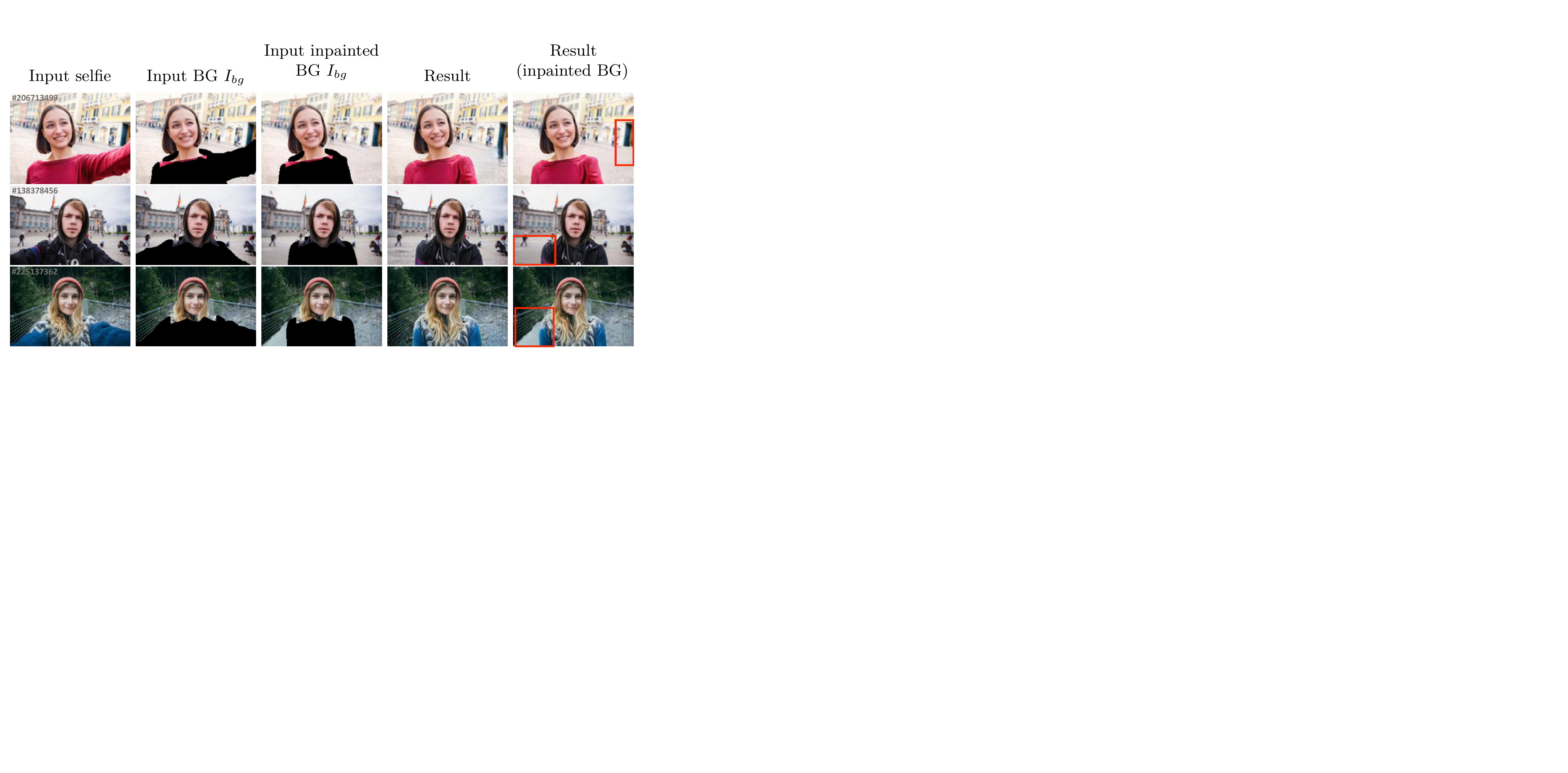}\\
\caption{Results of using inpaited background.}
\label{supp_fig:inpaintBG}
\end{figure}

In Fig.~\ref{supp_fig:inpaintBG}, we provide side-by-side result comparisons of our original pipeline and a slightly modified pipeline using off-the-shelf inpainting network~\cite{zeng2020high} to fill the dis-occluded background holes before feeding the inpainted background ($I_{bg}$ in Figure 6 of the main paper) into our composition stage. 
In particular, during inference, we use the pre-trained inpainting network to inpaint the holes $H=H_{selfie}$ (marked in black in second column of Fig.~\ref{supp_fig:inpaintBG}). Then we apply a matting algorithm~\cite{deepmatting} to the retrieved \new{neutral-pose} portrait to create a new hole $H=H_{neutral}$ (third column of Fig.~\ref{supp_fig:inpaintBG}) before feeding it together with the coordinate inpainting result ($I_{fg}$ in Figure 6 of the main paper) as input to our composition network. In Fig.~\ref{supp_fig:inpaintBG}, the fourth column shows the result of our original pipeline and the fifth column shows the result of our modified pipeline leveraging the inpainting network.

Theoretically, using a separate inpainting network to handle the background separately allows us to harvest the latest advances in image inpainting and focus our composition module exclusively on the synthesis of foreground details and foreground-background transitions. 
However, there are pros and cons in practice.
Pros: the pretrained inpainting network can help reduce the artifacts in the background holes and near the foreground boundaries. For example, in the top row of Fig.~\ref{supp_fig:inpaintBG}, the structure of the door on the right side of the image is better synthesized. In the top and middle rows, the artifacts near the arms are also reduced. Cons: occasionally the inpainting network could introduce small artifacts near the foreground boundary. For example, in the bottom row, the inpainting network introduced some grey regions on top of the girl's right shoulder.

\section{More comparisons}
\begin{figure*} [htp]
\centering

\begin{subfigure}[b]{.16\linewidth}
\caption*{Input selfie}
\vspace{-1mm}
\begin{overpic}[width=\linewidth]{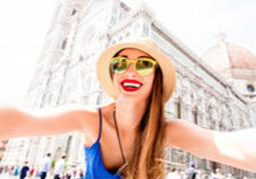}
\put(0,620) {\scalebox{0.65}{\id{119222256}}}
\end{overpic}
\end{subfigure}
\begin{subfigure}[b]{.16\linewidth}
\caption*{DPIG}
\vspace{-1mm}
\includegraphics[width=\linewidth]{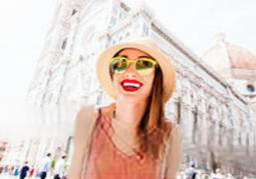}
\end{subfigure}
\begin{subfigure}[b]{.16\linewidth}
\caption*{VUNET}
\vspace{-1mm}
\includegraphics[width=\linewidth]{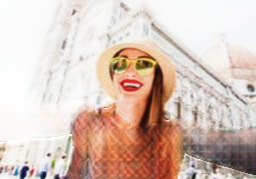}
\end{subfigure}
\begin{subfigure}[b]{.16\linewidth}
\caption*{PATN}
\vspace{-1mm}
\includegraphics[width=\linewidth]{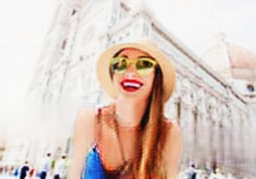}
\end{subfigure}
\begin{subfigure}[b]{.16\linewidth}
\caption*{Ours}
\vspace{-1mm}
\includegraphics[width=\linewidth]{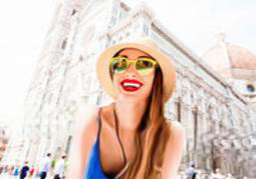}
\end{subfigure}
\begin{subfigure}[b]{.16\linewidth}
\caption*{Ours w/ inpainted BG}
\vspace{-1mm}
\includegraphics[width=\linewidth]{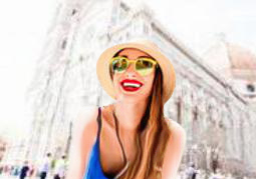}
\end{subfigure}


\begin{subfigure}[b]{.16\linewidth}
\begin{overpic}[width=\linewidth]{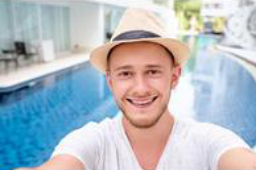}
\put(0,580) {\scalebox{0.65}{\id{166011716}}}
\end{overpic}
\end{subfigure}
\begin{subfigure}[b]{.16\linewidth}
\includegraphics[width=\linewidth]{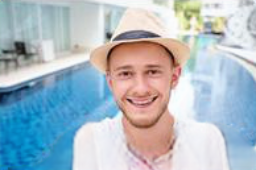}
\end{subfigure}
\begin{subfigure}[b]{.16\linewidth}
\includegraphics[width=\linewidth]{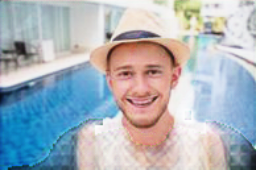}
\end{subfigure}
\begin{subfigure}[b]{.16\linewidth}
\includegraphics[width=\linewidth]{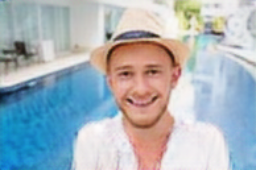}
\end{subfigure}
\begin{subfigure}[b]{.16\linewidth}
\includegraphics[width=\linewidth]{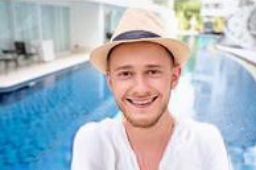}
\end{subfigure}
\begin{subfigure}[b]{.16\linewidth}
\includegraphics[width=\linewidth]{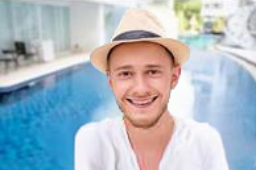}
\end{subfigure}

\begin{subfigure}[b]{.16\linewidth}
\begin{overpic}[width=\linewidth]{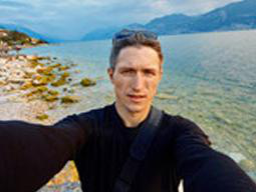}
\put(0,665) {\scalebox{0.65}{\id{96848570}}}
\end{overpic}
\end{subfigure}
\begin{subfigure}[b]{.16\linewidth}
\includegraphics[width=\linewidth]{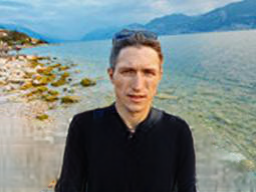}
\end{subfigure}
\begin{subfigure}[b]{.16\linewidth}
\includegraphics[width=\linewidth]{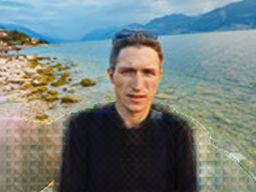}
\end{subfigure}
\begin{subfigure}[b]{.16\linewidth}
\includegraphics[width=\linewidth]{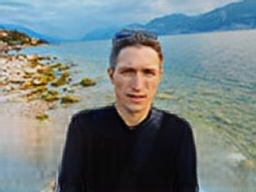}
\end{subfigure}
\begin{subfigure}[b]{.16\linewidth}
\includegraphics[width=\linewidth]{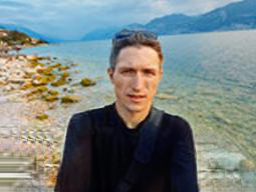}
\end{subfigure}
\begin{subfigure}[b]{.16\linewidth}
\includegraphics[width=\linewidth]{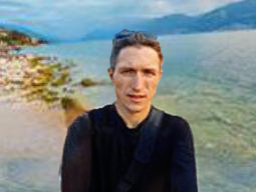}
\end{subfigure}

\begin{subfigure}[b]{.16\linewidth}
\begin{overpic}[width=\linewidth]{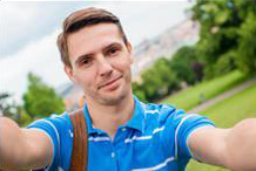}
\put(0,600) {\scalebox{0.65}{\id{116496273}}}
\end{overpic}
\end{subfigure}
\begin{subfigure}[b]{.16\linewidth}
\includegraphics[width=\linewidth]{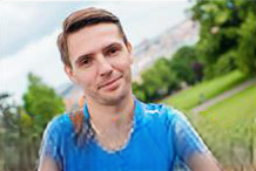}
\end{subfigure}
\begin{subfigure}[b]{.16\linewidth}
\includegraphics[width=\linewidth]{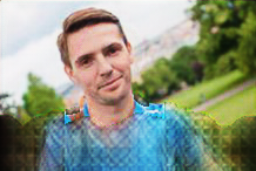}
\end{subfigure}
\begin{subfigure}[b]{.16\linewidth}
\includegraphics[width=\linewidth]{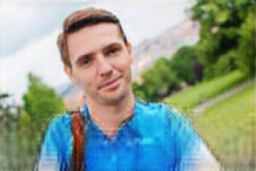}
\end{subfigure}
\begin{subfigure}[b]{.16\linewidth}
\includegraphics[width=\linewidth]{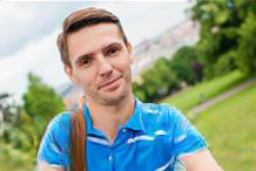}
\end{subfigure}
\begin{subfigure}[b]{.16\linewidth}
\includegraphics[width=\linewidth]{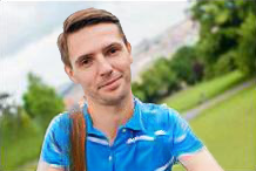}
\end{subfigure}

\caption{Comparisons with state-of-the-art methods.}
\label{supp_fig:compare}
\end{figure*}

In Fig.~\ref{supp_fig:compare}, we provide more comparisons between our approach and prior approaches, including VUNET~\cite{VUNET}. \textbf{Ours} is the result from our original pipeline where we manually picked the best one out of results using our top-5 retrieved poses. \textbf{Ours w/ inpainted BG} uses off-the-shelf inpainting network as described above. VUNET produces many artifacts in both body and background regions.

\section{Multi-modal results}
\begin{figure*} [htp]
\centering

\begin{subfigure}[b]{.16\linewidth}
\caption*{Input selfie}
\vspace{-1mm}
\begin{overpic}[width=\linewidth]{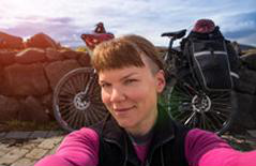}
\put(0,570) {\scalebox{0.65}{\id{101106916}}}
\end{overpic}
\end{subfigure}
\begin{subfigure}[b]{.16\linewidth}
\caption*{Top-1 result}
\vspace{-1mm}
\includegraphics[width=\linewidth]{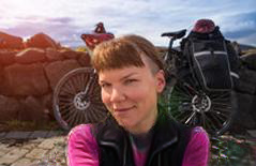}
\end{subfigure}
\begin{subfigure}[b]{.16\linewidth}
\caption*{Top-2 result}
\vspace{-1mm}
\includegraphics[width=\linewidth]{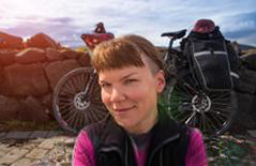}
\end{subfigure}
\begin{subfigure}[b]{.16\linewidth}
\caption*{Top-3 result}
\vspace{-1mm}
\includegraphics[width=\linewidth]{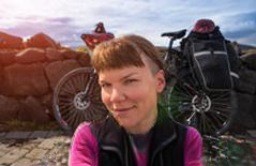}
\end{subfigure}
\begin{subfigure}[b]{.16\linewidth}
\caption*{Top-4 result}
\vspace{-1mm}
\includegraphics[width=\linewidth]{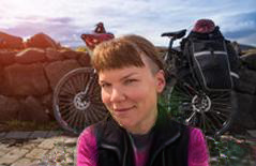}
\end{subfigure}
\begin{subfigure}[b]{.16\linewidth}
\caption*{Top-5 result}
\vspace{-1mm}
\includegraphics[width=\linewidth]{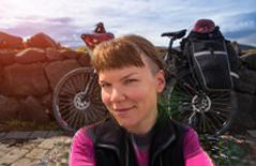}
\end{subfigure}

\begin{subfigure}[b]{.16\linewidth}
\begin{overpic}[width=\linewidth]{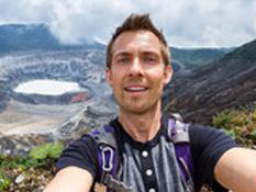}
\put(0,650) {\scalebox{0.65}{\id{120915150}}}
\end{overpic}
\end{subfigure}
\begin{subfigure}[b]{.16\linewidth}
\includegraphics[width=\linewidth]{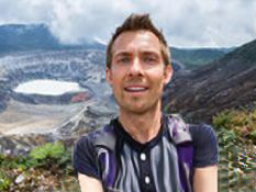}
\end{subfigure}
\begin{subfigure}[b]{.16\linewidth}
\includegraphics[width=\linewidth]{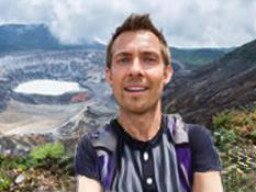}
\end{subfigure}
\begin{subfigure}[b]{.16\linewidth}
\includegraphics[width=\linewidth]{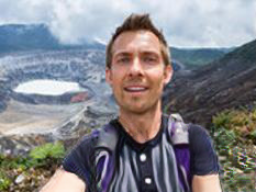}
\end{subfigure}
\begin{subfigure}[b]{.16\linewidth}
\includegraphics[width=\linewidth]{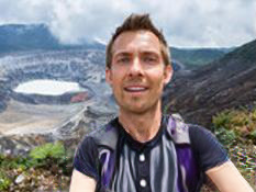}
\end{subfigure}
\begin{subfigure}[b]{.16\linewidth}
\includegraphics[width=\linewidth]{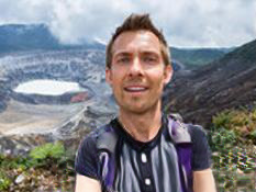}
\end{subfigure}

\begin{subfigure}[b]{.16\linewidth}
\begin{overpic}[width=\linewidth]{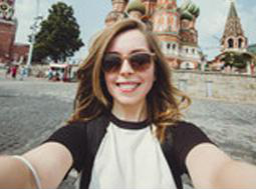}
\put(0,650) {\scalebox{0.65}{\id{133457041}}}
\end{overpic}
\end{subfigure}
\begin{subfigure}[b]{.16\linewidth}
\includegraphics[width=\linewidth]{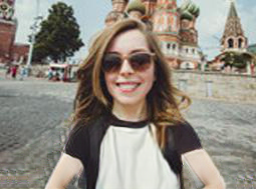}
\end{subfigure}
\begin{subfigure}[b]{.16\linewidth}
\includegraphics[width=\linewidth]{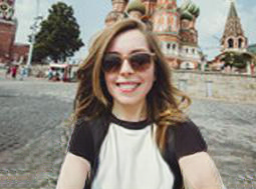}
\end{subfigure}
\begin{subfigure}[b]{.16\linewidth}
\includegraphics[width=\linewidth]{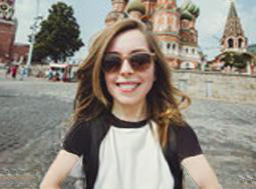}
\end{subfigure}
\begin{subfigure}[b]{.16\linewidth}
\includegraphics[width=\linewidth]{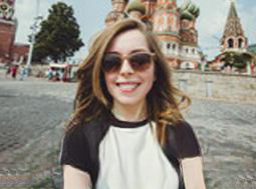}
\end{subfigure}
\begin{subfigure}[b]{.16\linewidth}
\includegraphics[width=\linewidth]{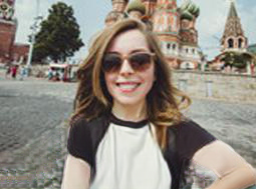}
\end{subfigure}


\begin{subfigure}[b]{.16\linewidth}
\begin{overpic}[width=\linewidth]{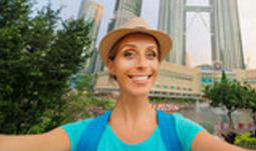}
\put(0,510) {\scalebox{0.65}{\id{135312945}}}
\end{overpic}
\end{subfigure}
\begin{subfigure}[b]{.16\linewidth}
\includegraphics[width=\linewidth]{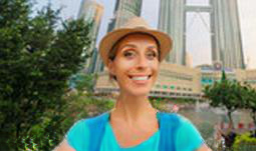}
\end{subfigure}
\begin{subfigure}[b]{.16\linewidth}
\includegraphics[width=\linewidth]{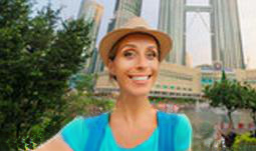}
\end{subfigure}
\begin{subfigure}[b]{.16\linewidth}
\includegraphics[width=\linewidth]{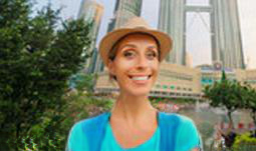}
\end{subfigure}
\begin{subfigure}[b]{.16\linewidth}
\includegraphics[width=\linewidth]{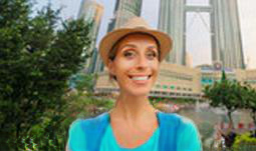}
\end{subfigure}
\begin{subfigure}[b]{.16\linewidth}
\includegraphics[width=\linewidth]{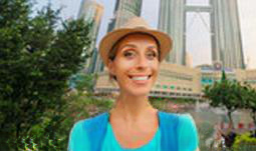}
\end{subfigure}

\begin{subfigure}[b]{.16\linewidth}
\begin{overpic}[width=\linewidth]{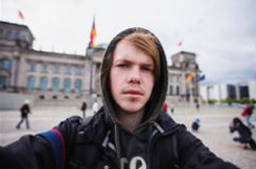}
\put(0,570) {\scalebox{0.65}{\id{138378456}}}
\end{overpic}
\end{subfigure}
\begin{subfigure}[b]{.16\linewidth}
\includegraphics[width=\linewidth]{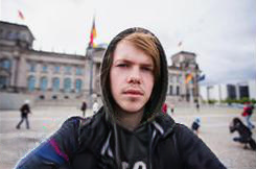}
\end{subfigure}
\begin{subfigure}[b]{.16\linewidth}
\includegraphics[width=\linewidth]{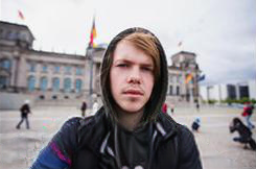}
\end{subfigure}
\begin{subfigure}[b]{.16\linewidth}
\includegraphics[width=\linewidth]{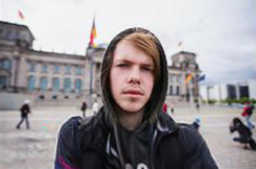}
\end{subfigure}
\begin{subfigure}[b]{.16\linewidth}
\includegraphics[width=\linewidth]{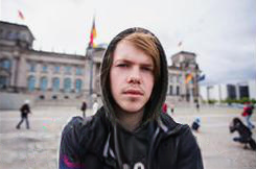}
\end{subfigure}
\begin{subfigure}[b]{.16\linewidth}
\includegraphics[width=\linewidth]{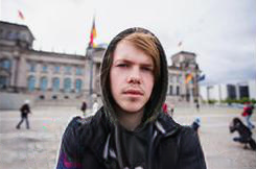}
\end{subfigure}


\begin{subfigure}[b]{.16\linewidth}
\begin{overpic}[width=\linewidth]{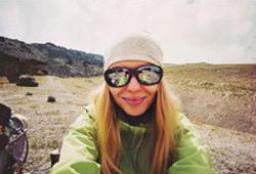}
\put(0,620) {\scalebox{0.65}{\id{182146016}}}
\end{overpic}
\end{subfigure}
\begin{subfigure}[b]{.16\linewidth}
\includegraphics[width=\linewidth]{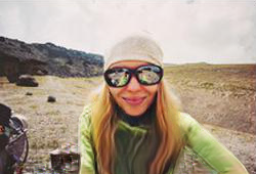}
\end{subfigure}
\begin{subfigure}[b]{.16\linewidth}
\includegraphics[width=\linewidth]{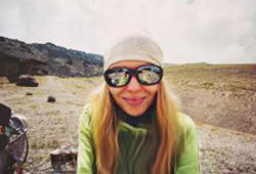}
\end{subfigure}
\begin{subfigure}[b]{.16\linewidth}
\includegraphics[width=\linewidth]{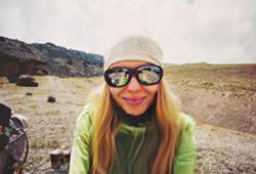}
\end{subfigure}
\begin{subfigure}[b]{.16\linewidth}
\includegraphics[width=\linewidth]{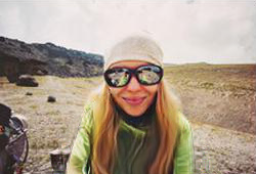}
\end{subfigure}
\begin{subfigure}[b]{.16\linewidth}
\includegraphics[width=\linewidth]{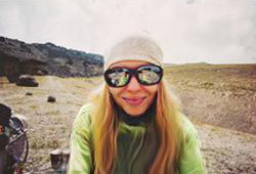}
\end{subfigure}


\begin{subfigure}[b]{.16\linewidth}
\begin{overpic}[width=\linewidth]{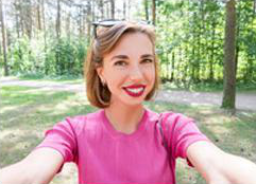}
\put(0,630) {\scalebox{0.65}{\id{212727509}}}
\end{overpic}
\end{subfigure}
\begin{subfigure}[b]{.16\linewidth}
\includegraphics[width=\linewidth]{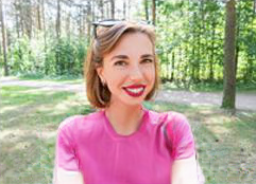}
\end{subfigure}
\begin{subfigure}[b]{.16\linewidth}
\includegraphics[width=\linewidth]{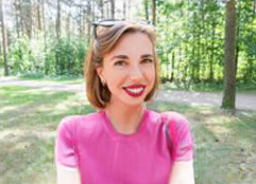}
\end{subfigure}
\begin{subfigure}[b]{.16\linewidth}
\includegraphics[width=\linewidth]{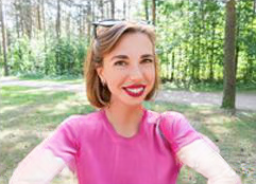}
\end{subfigure}
\begin{subfigure}[b]{.16\linewidth}
\includegraphics[width=\linewidth]{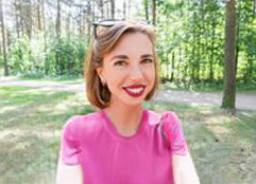}
\end{subfigure}
\begin{subfigure}[b]{.16\linewidth}
\includegraphics[width=\linewidth]{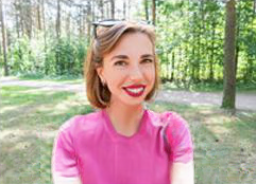}
\end{subfigure}

\begin{subfigure}[b]{.16\linewidth}
\begin{overpic}[width=\linewidth]{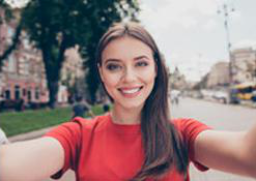}
\put(0,630) {\scalebox{0.65}{\id{218021773}}}
\end{overpic}
\end{subfigure}
\begin{subfigure}[b]{.16\linewidth}
\includegraphics[width=\linewidth]{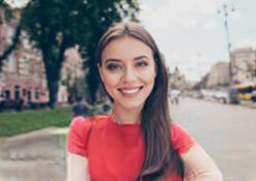}
\end{subfigure}
\begin{subfigure}[b]{.16\linewidth}
\includegraphics[width=\linewidth]{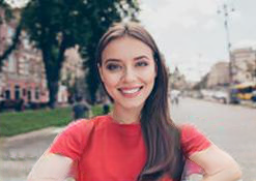}
\end{subfigure}
\begin{subfigure}[b]{.16\linewidth}
\includegraphics[width=\linewidth]{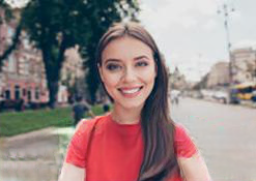}
\end{subfigure}
\begin{subfigure}[b]{.16\linewidth}
\includegraphics[width=\linewidth]{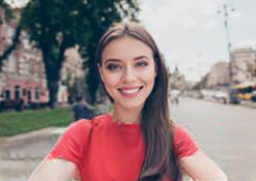}
\end{subfigure}
\begin{subfigure}[b]{.16\linewidth}
\includegraphics[width=\linewidth]{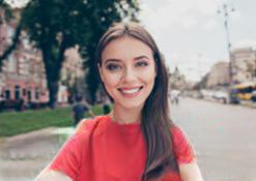}
\end{subfigure}


\begin{subfigure}[b]{.16\linewidth}
\begin{overpic}[width=\linewidth]{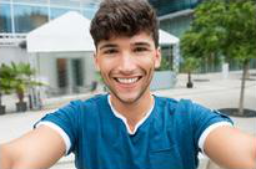}
\put(0,580) {\scalebox{0.65}{\id{92379867}}}
\end{overpic}
\end{subfigure}
\begin{subfigure}[b]{.16\linewidth}
\includegraphics[width=\linewidth]{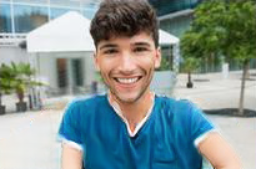}
\end{subfigure}
\begin{subfigure}[b]{.16\linewidth}
\includegraphics[width=\linewidth]{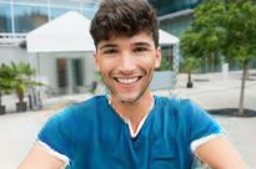}
\end{subfigure}
\begin{subfigure}[b]{.16\linewidth}
\includegraphics[width=\linewidth]{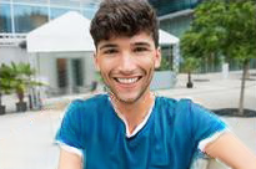}
\end{subfigure}
\begin{subfigure}[b]{.16\linewidth}
\includegraphics[width=\linewidth]{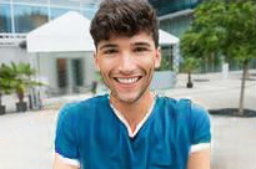}
\end{subfigure}
\begin{subfigure}[b]{.16\linewidth}
\includegraphics[width=\linewidth]{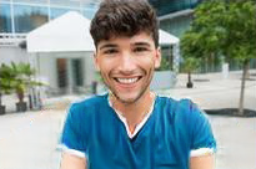}
\end{subfigure}

\caption{Top-k results. 1st column: the input selfie image. 2-6th columns: the top-k unselfie results.}
\label{supp_fig:topk_result}
\end{figure*}

As mentioned in the main paper, our nearest pose search module can generate multiple output variations based on the same input selfie. Fig.~\ref{supp_fig:topk_result} provides top5 results for every input selfie. Most of the top5 results have consistent quality with each other. This gives users the freedom to choose the best pose they prefer.

\section{Implementation details}
\label{supp_sec:details}

\myparagraph{Image alignment.}
As mentioned in the main paper, we align the image and pose into the center part of a 256$\times$256 resolution canvas. Likewise, the coordinate-map and texture-map are also in 256$\times$256 resolution.
To align the image and pose, we use two shoulder points whose locations are at (63,133) and (92,133) on the 256$\times$256 coordinate-map. After obtaining the coordinates of the two shoulder points from the coordinate-map, we calculate the scale and translation factors for image and pose alignment by aligning the shoulder points to (112,128) and (143,128) on the 256$\times$256 image.

\myparagraph{Hyper-parameters and miscellaneous details.}
For model optimization, we use the Adam optimizer with $\beta_1=0.5$, $\beta_2=0.999$. 
$G1$ is trained with a minibatch of size 10 for 70k iterations with initial learning rate of $0.0001$. $G2$ is trained with a minibatch of size 2 for 400k iterations with initial learning rate of $0.00002$.
The loss weights are set to  $\lambda_1=2$, $\lambda_2=10$, $\lambda_3=\lambda_4=\lambda_5=10$,
We use three types of data augmentation during training: 1) left-right image flip; 2) background replacement through foreground mask estimation~\cite{deepmatting}; 3) random paired selfie selection among top-40 retrieved results.
As to the output, we mask out the generated pixels in the invalid region $M$ which denotes the invalid region of the image caused by the alignment step. Therefore, the final output can be formulated as follows,
\begin{equation}
    I_{out}=(I_{G_2} A_{G_2} + I_{bg}(1-A_{G_2})) (1-M).
    \label{supp_eq:I_out}  
\end{equation}
The Image2UV (I2UV) mapping is implemented via a lookup table follows~\cite{alldieck2019tex2shape}

\myparagraph{Improvement for DPIG~\cite{DPIG} and VUNET~\cite{VUNET}}
As mentioned in the main paper, we made various improvements for DPIG~\cite{DPIG} to produce comparable results to ours, because the DPIG model does not converge during training when directly applied to our task. One possible reason is that the background and human appearance in our data contain a lot of variations which are very hard to model in the latent space. 
For fairer comparison, we improve DPIG in several ways, including adding $I_{bg}$ as input to the decoder, adding perceptual loss~\cite{LPIPS}, using resnet-based PatchGAN discriminator with LSGAN loss~\cite{poseTransfer}.
We also improve VUNET by adding $I_{bg}$ as input to U-net encoder and adding $L_1$ loss to stabilize training. We also tried adding adversarial loss but observe little improvement.

\myparagraph{Network architectures.}
As to our inpainting network architecture, we use the same network as that of~\cite{CBI_coordInpaint}, except that our input contains 5 channels including $C_{src}$ (2 channels) and $T_{src}$ (3 channels).
Our composition network consists of source encoder branch, target encoder branch, and decoder as shown in Tab.~\ref{supp_tab:network}. 
The notations $src\_blkN, N=1,...,3$, $tgt\_blkN, N=1,...,3$, $res\_blkN, N =
1,...,6$ corresponds to a block with gated convolution layer proposed in~\cite{yu2019free} followed by group normalization~\cite{wu2018group} (group number = 32) and Leaky ReLU (slope = 0.01).
%


\begin{table}
\setlength{\tabcolsep}{5.5pt} 
\centering
\footnotesize
\caption{The composition network architecture.} 
\begin{adjustbox}{width=1\columnwidth,center}
\renewcommand{\arraystretch}{1.8}
\begin{tabular}
{@{\extracolsep{\fill}} l l l l l}
\toprule 
Layer & \makecell{Filters/Stride\\ (Dilation)} & Input & Input Size  & Output Size     \\
\midrule[0.6pt]	
\multicolumn{5}{l}{Source encoder branch} \\
\midrule[0.6pt]	
src\_blk1 & 5 x 5 / 1 (1) & [$P_{src}$, $I_{fg}$] & 6 x H x W & 256 x H x W     \\
src\_blk2 & 3 x 3 / 1 (1) & src\_blk1 & 256 x H x W & 256 x H x W     \\
src\_blk3 & 3 x 3 / 1 (2) & src\_blk2 & 256 x H x W & 256 x H x W     \\
\midrule[0.6pt]	
\multicolumn{5}{l}{Target encoder branch} \\
\midrule[0.6pt]	
tgt\_blk1 & 5 x 5 / 1 (1) & \makecell[l]{[$P_{tgt}$, $I_{bg}$, $I_{G1}$, M]} & 10 x H x W  & 256 x H x W \\
tgt\_blk2 & 3 x 3 / 2 (1) & tgt\_blk1 & 256 x H x W & 256 x $\frac{H}{2}$ x $\frac{W}{2}$ \\
tgt\_blk3 & 3 x 3 / 1 (1) & tgt\_blk2 & 256 x $\frac{H}{2}$ x $\frac{W}{2}$ & 256 x $\frac{H}{4}$ x $\frac{W}{4}$\\
res\_blk1 & 3 x 3 / 1 (1) & tgt\_blk3 & 256 x $\frac{H}{4}$ x $\frac{W}{4}$ & 256 x $\frac{H}{4}$ x $\frac{W}{4}$\\
res\_blk2 & 3 x 3 / 1 (1) & \makecell[l]{[res\_blk1 + tgt\_blk3]} & 256 x $\frac{H}{4}$ x $\frac{W}{4}$ & 256 x $\frac{H}{4}$ x $\frac{W}{4}$\\
res\_blk3 & 3 x 3 / 1 (1) & \makecell[l]{[res\_blk1 + res\_blk2]} & 256 x $\frac{H}{4}$ x $\frac{W}{4}$ & 256 x $\frac{H}{4}$ x $\frac{W}{4}$\\
res\_blk4 & 3 x 3 / 1 (1) & \makecell[l]{[res\_blk2 + res\_blk3]} & 256 x $\frac{H}{4}$ x $\frac{W}{4}$ & 256 x $\frac{H}{4}$ x $\frac{W}{4}$\\
res\_blk5 & 3 x 3 / 1 (1) & \makecell[l]{[res\_blk3 + res\_blk4]} & 256 x $\frac{H}{4}$ x $\frac{W}{4}$ & 256 x $\frac{H}{4}$ x $\frac{W}{4}$\\
res\_blk6 & 3 x 3 / 1 (1) & \makecell[l]{[res\_blk4 + res\_blk5]} & 256 x $\frac{H}{4}$ x $\frac{W}{4}$ & 256 x $\frac{H}{4}$ x $\frac{W}{4}$\\
\midrule[0.6pt]	
\multicolumn{5}{l}{Decoder} \\
\midrule[0.6pt]	
dec\_blk1 & 3 x 3 / 1 (1) & \makecell[l]{[res\_blk5 + res\_blk6, \\warp(src\_blk3,E), \\tgt\_blk3]} & 768 x $\frac{H}{4}$ x $\frac{W}{4}$ & 256 x $\frac{H}{4}$ x $\frac{W}{4}$\\
upsample1 & --- & dec\_blk1 & 256 x $\frac{H}{4}$ x $\frac{W}{4}$ & 256 x $\frac{H}{2}$ x $\frac{W}{2}$\\
dec\_blk2 & 3 x 3 / 1 (1) & \makecell[l]{[upsample1, \\warp(src\_blk2,E), \\tgt\_blk2]} & 768 x $\frac{H}{2}$ x $\frac{W}{2}$ & 256 x $\frac{H}{2}$ x $\frac{W}{2}$\\
upsample2 & --- & dec\_blk2 & 256 x $\frac{H}{2}$ x $\frac{W}{2}$ & 256 x $H$ x $W$\\
dec\_blk3 & 3 x 3 / 1 (1) & \makecell[l]{[upsample2, \\warp(src\_blk1,E), \\tgt\_blk1]} & 768 x $H$ x $W$ & 256 x $H$ x $W$\\
dec\_blk4 & 3 x 3 / 1 (1) & dec\_blk3 & 256 x $H$ x $W$ & 6 x $H$ x $W$\\
tanh & --- & dec\_blk4 & 6 x $H$ x $W$ & 6 x $H$ x $W$\\
\bottomrule 
\end{tabular}
\end{adjustbox}
\label{supp_tab:network}
\end{table}

\clearpage
\section{Attribution:}
\new{Selfie photo owners: \#206713499-Paolese, \#138378456-iiievgeniy, \#225137362-BublikHaus, \#119222256-rh2010, \#166011716-luengo\_ua, \#96848570-vitaliymateha, \#116496273-travnikovstudio, \#101106916-lkoimages, \#120915150-wollertz, \#133457041-ilovemayorova, \#135312945-luengo\_ua, \#182146016-EVERST, \#212727509-Photocatcher, \#218021773-deagreez, \#92379867-Rido -- stock.adobe.com.
}


\clearpage
\bibliographystyle{splncs04}
\bibliography{eccv20} 

\end{document}